\documentclass[conference]{IEEEtran}
\IEEEoverridecommandlockouts
\usepackage[accsupp]{axessibility}  
\usepackage{cite}
\usepackage{amsmath}
\usepackage{amsfonts}
\usepackage{amssymb}
\usepackage{graphicx}
\usepackage{textcomp}
\usepackage{xcolor}
\usepackage{comment}
\usepackage [latin1]{inputenc}

\usepackage{multirow}
\usepackage{algorithm}
\usepackage{algorithmicx}
\usepackage{algpseudocode}
\usepackage{url}
\usepackage{soul}
\usepackage{balance}

\usepackage{amssymb}
\usepackage{booktabs}
\usepackage[percent]{overpic}

\usepackage{xcolor}
\usepackage{makecell}

\def\BibTeX{{\rm B\kern-.05em{\sc i\kern-.025em b}\kern-.08em
    T\kern-.1667em\lower.7ex\hbox{E}\kern-.125emX}}
\begin{document}

\title{Multi-scale Progressive Feature Embedding for Accurate
NIR-to-RGB Spectral Domain Translation}
\author{Anonymous Author(s)\\Submission ID: 175}

\author{\IEEEauthorblockN{Xingxing Yang}
\IEEEauthorblockA{\textit{Department of Computer Science} \\
\textit{Hong Kong Baptist University}\\
Hong Kong SAR, China \\
csxxyang@comp.hkbu.edu.hk}
\and
\IEEEauthorblockN{Jie Chen}
\IEEEauthorblockA{\textit{Department of Computer Science} \\
\textit{Hong Kong Baptist University}\\
Hong Kong SAR, China \\
chenjie@comp.hkbu.edu.hk}
\and
\IEEEauthorblockN{Zaifeng Yang}
\IEEEauthorblockA{\textit{Institute of High Performance Computing} \\
\textit{Agency for Science, Technology and Research}\\
Singapore \\
yang\_zaifeng@ihpc.a-star.edu.sg}
}

\maketitle

\begin{abstract}
NIR-to-RGB spectral domain translation is a challenging task due to the mapping ambiguities and existing methods show limited learning capacities. To address these challenges, we propose to colorize NIR images via a multi-scale progressive feature embedding network (MPFNet), with the guidance of grayscale image colorization.
Specifically, we first introduce a domain translation module that translates NIR source images into the grayscale target domain. By incorporating a progressive training strategy, the statistical and semantic knowledge from both task domains are efficiently aligned with a series of pixel-/feature-level consistency constraints. Besides, a multi-scale progressive feature embedding network is designed to improve learning capabilities. Experiments show that our MPFNet outperforms state-of-the-art counterparts by \textbf{2.55dB} in the NIR-to-RGB spectral domain translation task in terms of PSNR.
\end{abstract}


\section{Introduction}\label{introduction}

\IEEEPARstart{N}{ear-infrared} (NIR) imaging systems can capture unique spectral reflectance details, which are widely used in night-time video surveillance \cite{christnacher2018portable}, object detection, material analysis \cite{liu2019improved} and remote sensing systems \cite{protopapadakis2021stacked}.
NIR domain information (780 nm to 2500 nm), though having all the unique application values, is neither natural nor efficient for both human and computer vision systems to explore. Consequently, NIR-to-RGB spectral domain translation has become a valuable research topic. 
Recent development in deep learning has brought great advancement to image translation tasks, like grayscale image colorization \cite{Deshpande_2015_ICCV, cheng2015deep, 9257445}. However, the progress of NIR-to-RGB spectral domain translation \cite{9301791, 9301839, 9301787} lags behind, and the reasons are analyzed as follows:

\begin{figure}
    \centering
  { 
      \includegraphics[width=1\linewidth]{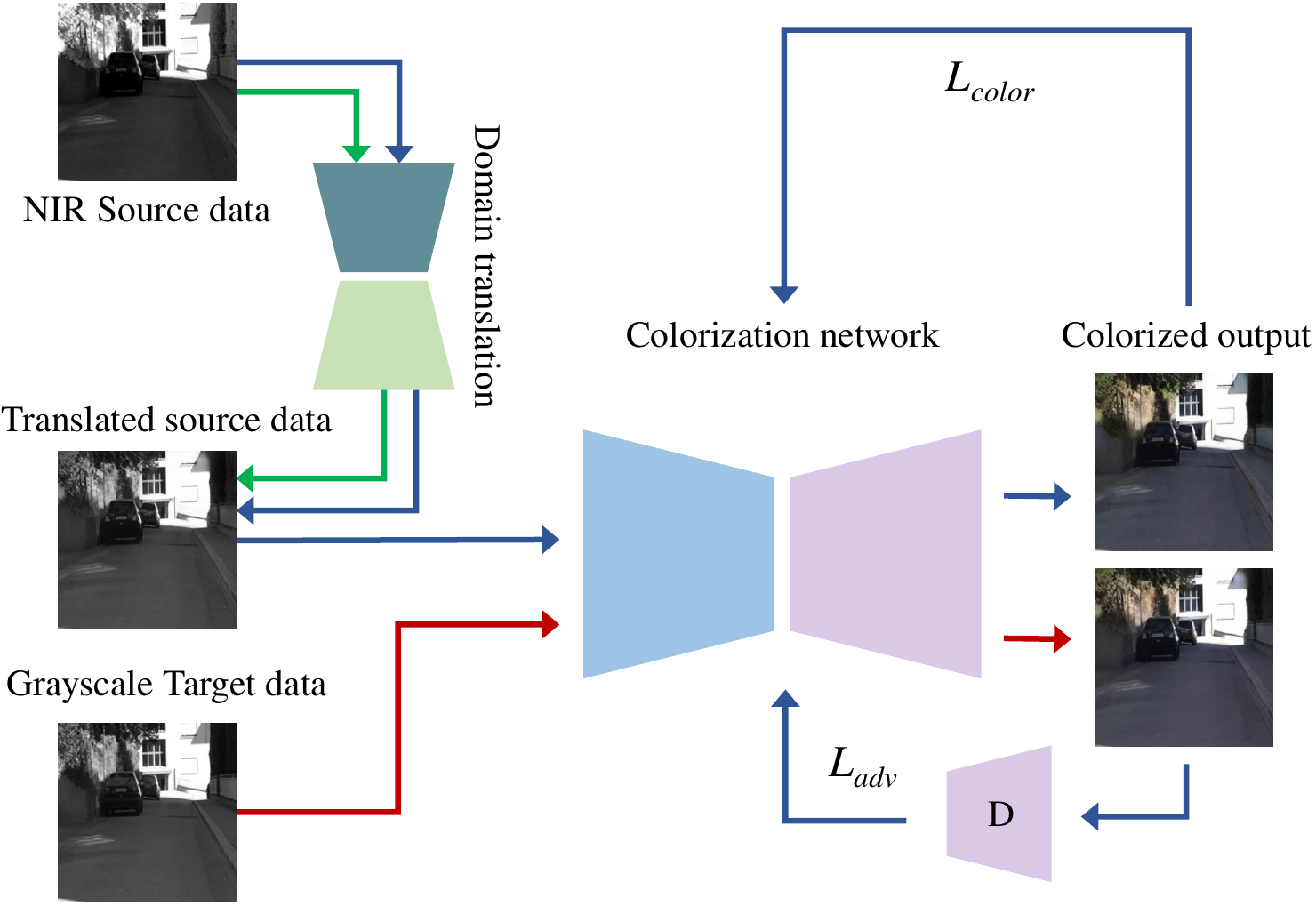}}
  \caption{The overview of the spectrum translation. NIR source images are firstly translated into the grayscale domain by a domain translation module, and then colorized by a pre-trained grayscale image colorization network. The green line indicates the flow of the pre-training process for the domain translation module; the red line indicates the flow of the pre-training process for the grayscale image colorization network and the blue line indicates the flow of the fine-tuning process of NIR inputs. By colorization loss ($L_\text{color}$) of the colorized translated source data and RGB ground truths, the model learns implicit mapping relationships. By adversarial loss ($L_\text{adv}$), the model reduces the distribution gap between NIR source data and grayscale target data in feature space.}
    \label{first view} 
\end{figure}

(\textbf{\romannumeral1}) \textbf{Mapping Ambiguity.} 
NIR-to-RGB spectral domain translation\cite{10.1007/978-3-319-61578-3_16, 8014766} is intrinsically a much more challenging task compared with grayscale colorization due to the mapping ambiguity introduced by the domain gap between the non-overlapping spectral bands, which requires estimation of both luminance and chrominance values (while grayscale image colorization requires only the estimation of the latter).
This causes conventional image-to-image translation paradigms using given mapping ground-truth RGB images as supervision to be prone to produce monotonous, if not erroneous, predictions as the optimization process will push the prediction to approximate the statistical average \cite{9301788}. 
We find most existing methods tend to overlook this aspect \cite{9301839, 9301787, 9301788}.

(\textbf{\romannumeral2}) \textbf{Limited Learning Capability.} 
Existing methods (\cite{liang2021improved}, \cite{suarez2017infrared}, \cite{dong2018infrared}, \cite{9301788}) mostly stack convolutional layers to perform end-to-end prediction in a supervised manner. Some unsupervised \cite{cvpr2019_cycleGAN_NIR} and semi-supervised (\cite{sun2019nir}, \cite{9301791}) methods even adopt CycleGAN \cite{Conventional_CycleGAN} to translate the NIR domain to the RGB domain. However, all these methods produce unsatisfactory results because aligning the spectrum discrepancy is challenging.

\begin{figure*}[t]
    \centering
  { 
      \includegraphics[width=0.8\linewidth]{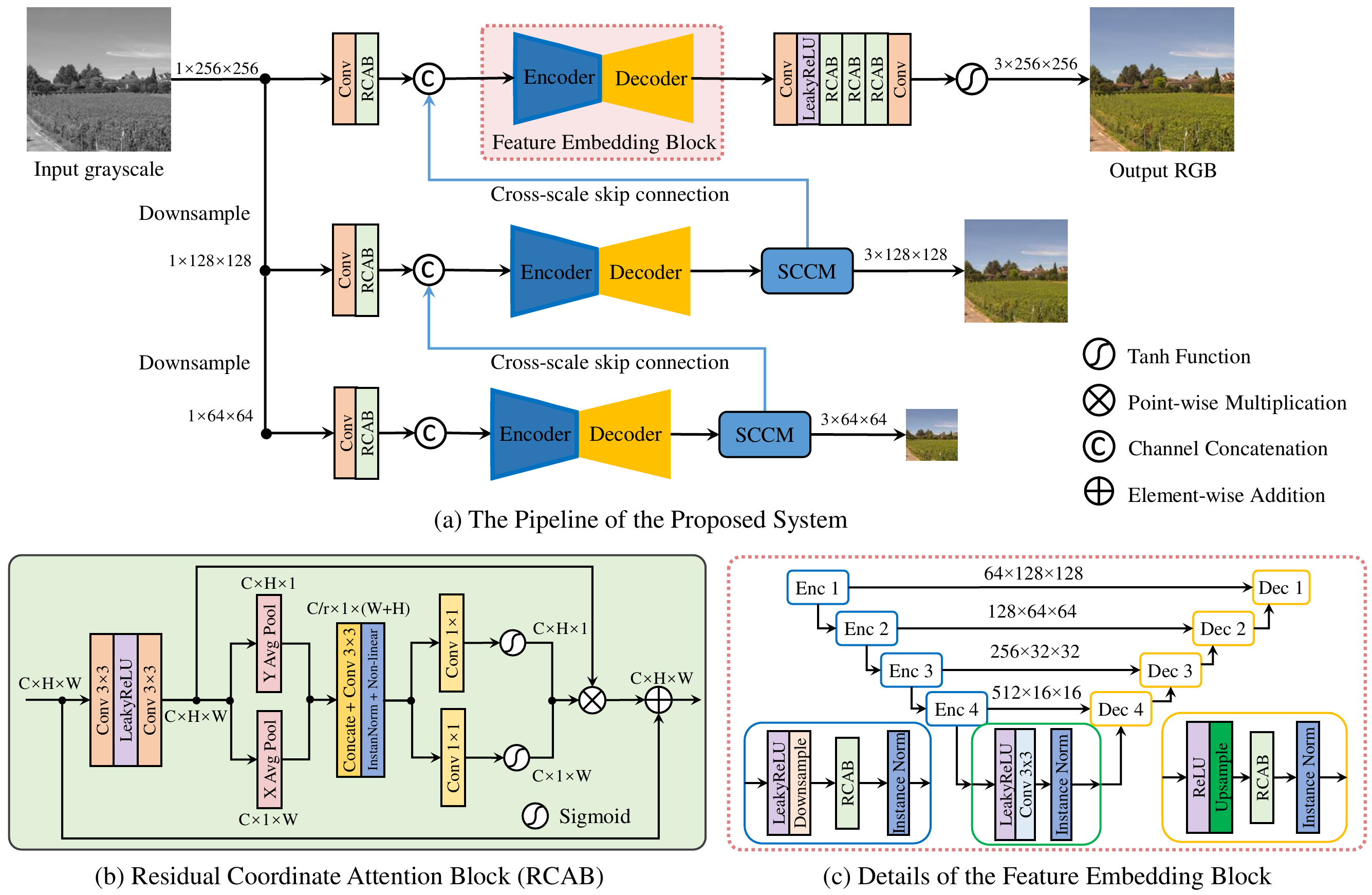}}
  \caption{(a) Proposed multi-scale progressive feature embedding network (MPFNet). The SCCM is inspired by \cite{mehri2021mprnet}. (b) and (c) illustrate structural details of the embedded Residual Coordinate Attention Block (RCAB) and the Feature Embedding Block (FEB), respectively.}
    \label{MPFNet} 
\end{figure*} 

\begin{figure*}
    \centering
  { 
      \includegraphics[width=1\linewidth]{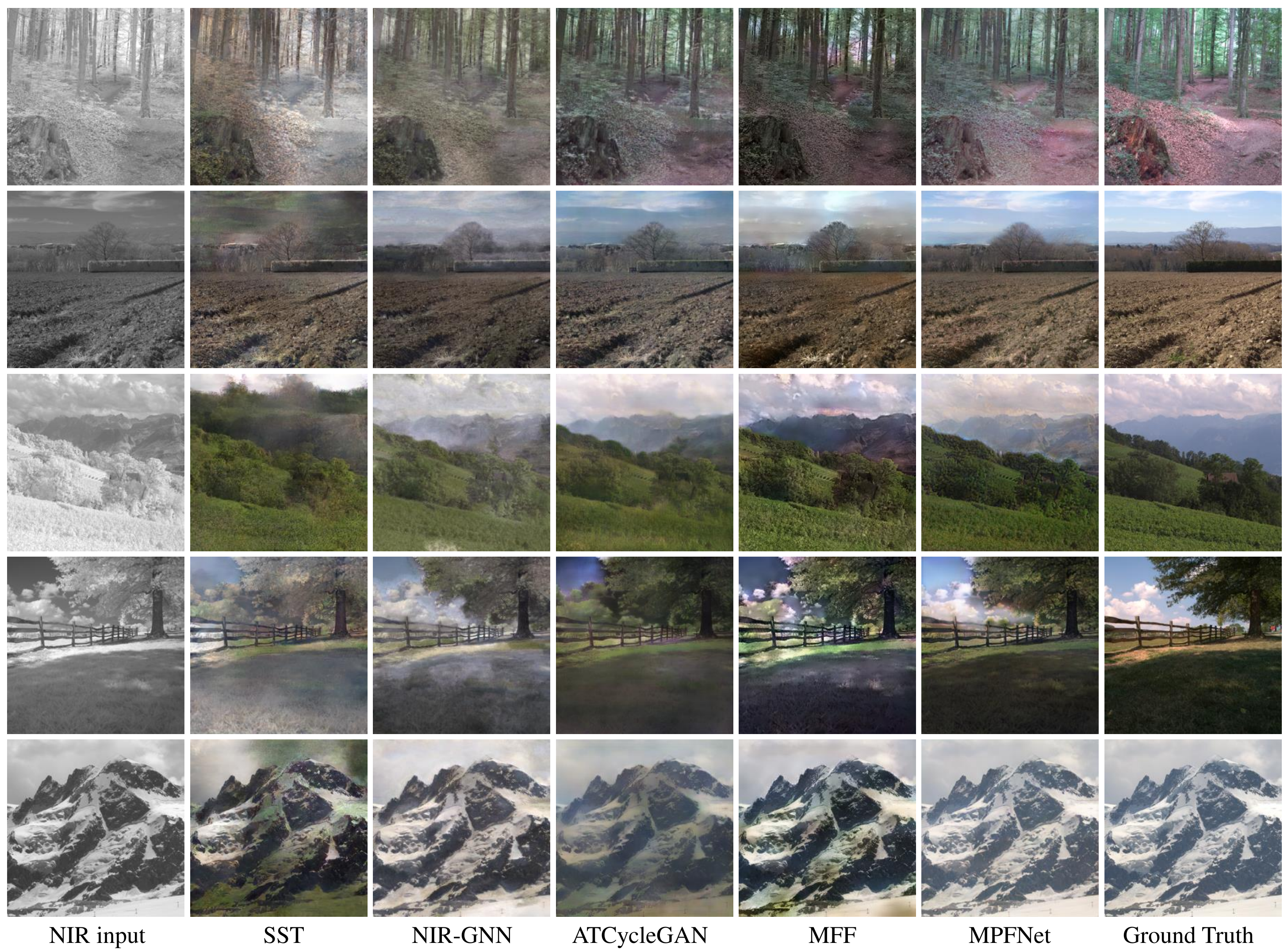}}
  \caption{Visual comparison among different NIR-to-RGB spectral domain translation methods. Images from left to right correspond to the NIR input images, and colorized results of SST \cite{9301788}, NIR-GNN \cite{9301839}, ATCycleGAN \cite{9301791}, MFF \cite{9301787}, our proposed MPFNet and RGB ground-truth images.}
    \label{result_gragh_1} 
\end{figure*}

Fortunately, learning-based grayscale image colorization is free of the above-mentioned two issues. As such, in this study, we propose to colorize NIR images with the guidance of grayscale image colorization, which is comprised of a domain translation module between NIR and grayscale images and a colorization module, as shown in Fig. \ref{first view}. Specifically, our framework first leverages a domain translation module that translates NIR source images into the grayscale target domain. 
For the colorization module, the colorization network is pre-trained by grayscale target images and subsequently fine-tuned by translated grayscale images with a series of pixel-level and feature-level consistency constraints to learn and fuse the statistical and semantic knowledge from both task domains.
To improve the learning capacity, a multi-scale progressive feature embedding network (MPFNet) is designed.

\section{Proposed Approach}\label{section_aproach}

\textbf{Method Overview.}
As shown in Fig. \ref{first view}, the proposed framework comprises a domain translation module and a colorization module. 
The domain translation module translates NIR source images into the grayscale target domain, which generates plausible grayscale images $X_{N2G}$ from the NIR domain.
The colorization module $F_G$ first translates grayscale target images $X_G$ into the RGB domain in the pre-training stage, and then translates plausible grayscale images $X_{N2G}$ into the RGB domain in the fine-tuning stage.
By incorporating the multi-scale feature embedding network as the backbone of both domain translation module and colorization module, the statistical and semantic knowledge from both NIR and grayscale domains are efficiently fused with a series of pixel-/feature-level consistency constraints.

\subsection{Multi-scale Feature Embedding Network}
Direct domain translation from NIR to RGB in a single path, as done in previous works in \cite{8014766, 9025662, 9301791}, produces unstable results both visually and semantically (illustrated in Fig. \ref{result_gragh_1}). 
In contrast, we propose a multi-scale encoder-decoder architecture for both NIR-to-grayscale domain translation and grayscale-to-RGB colorization. A schematic diagram of the system is shown in Fig. \ref{MPFNet}(a). 
The system breaks down the challenging domain mapping problem into sub-tasks with each focusing on a different resolution. 
At each scale, an encoder-decoder feature embedding block (FEB) is designed to learn contextual features (Fig. \ref{MPFNet}(c)), followed by a supervised color-consistency module (SCCM)\cite{mehri2021mprnet} which generates predictions with the supervision of ground-truth RGB images. 
Cross-scale skip connections are designed to propagate and fuse contextual details from lower resolutions to higher ones. 
To improve the learning capacity and highlight spatial image details, a residual coordinate attention block (RCAB) is introduced (Fig. \ref{MPFNet}(b)) and embedded into FEBs. 

Specifically, as shown in Fig. \ref{MPFNet} (c), 
FEBs adopt UNet \cite{yang2021attention} as the backbone to embed both contextual and textural features crucial for spectrum translation. 
Each encoder block consists of a leaky ReLU activation module and a stride-2 $4\times4$ convolution layer to downsample the feature map. Then, the residual coordinate attention block is designed to explore the spatial and channel feature correlation, followed by an instance normalization layer \cite{Huang_2017_ICCV} for promoting color style diversity \cite{jing2020dynamic}. 
The decoder block resembles the encoder but with a stride-2 $4\times4$ transpose convolution layer for feature upsampling.
The bottle-neck block contains only a $3\times3$ convolution layer without RCAB since the feature map is coarse and the attention mechanism becomes redundant. 
Different from \cite{Hu_2018_CVPR} and \cite{Woo_2018_ECCV} which perform feature re-weighting respectively in spatial and channel dimensions, as shown in Fig. \ref{MPFNet}(b), RCAB simultaneously captures both channel correlation and accurate position-sensitive information (via $x$ and $y$-axis average pooling).

\subsection{Objective Functions}
Our objective contains two terms: Domain Translation Losses to match the distribution of the NIR and grayscale domain; and Colorization Losses to map the functions from the grayscale domain to the RGB domain.

\textbf{Domain Translation Losses.}
Considering that we have both paired NIR-grayscale images and unpaired grayscale images in our dataset, we adopt a CycleGAN\cite{cvpr2019_cycleGAN_NIR} paradigm to train the domain translation module. The domain translation loss is defined as:

\begin{footnotesize}
\begin{align}
\label{domain translation}
\nonumber 
 & \mathcal{L}_{\text{tran}} =\mathcal{L}_{\text{GAN}}^{\text{img}}(X_{\text{N}},X_{\text{G}},D_{\text{N}}^{\text{img}},G_{\text{G2N}})+\mathcal{L}_{\text{GAN}}^{\text{img}}(X_{\text{N}},X_{\text{G}},D_{\text{G}}^{\text{img}},G_{\text{N2G}})\\ &+\lambda_{1} \mathcal{L}_{\text{cyc}}+\lambda_{2} \mathcal{L}_{\text{idt}},
\end{align}
\end{footnotesize}
where the $\mathcal{L}_{\text{GAN}}^{\text{img}}$, $ \mathcal{L}_{\text{cyc}}$ and $\mathcal{L}_{\text{idt}}$ are all defined in \cite{cvpr2019_cycleGAN_NIR}. $\lambda_{1}$ and $\lambda_{2}$ are hyperparameters, which we empirically set
both as 1.

\textbf{Colorization Losses.}
We employ a $L_{\text{mix}}$ loss \cite{zhao2016loss} at any given scale $s=1,2,..., S$ that combines both SSIM loss and L1 loss to formulate a supervised consistency constraint on both pixel and feature levels, which consists of two parts: during the pre-training stage, we only use original target images ($X_\text{G}$) to train the colorization module; while during the fine-tuning stage, we use  translated target images ($X_{\text{N2G}}$) as inputs, which are defined as:

\begin{footnotesize}
\begin{align}
\label{pair loss_1}
\mathcal{L}_{\text{pp}}& =\sum_{s=1}^{3}L_{\text{mix}}(Y_{\text{G}}^{s}, F_{\text{G}}^{s}\left(X_{\text{G}}\right)),\\
\label{pair loss_2}
\mathcal{L}_{\text{pf}}& =\sum_{s=1}^{3}L_{\text{mix}}(Y_{\text{N}}^{s}, F_{\text{G}}^{s}\left(X_{\text{N2G}}\right)),
\end{align}
\end{footnotesize} 
where $Y_\text{G}$ and $Y_\text{N}$ denote ground truths of grayscale images and NIR images, respectively. 

Additionally, in the fine-tuning stage, we further introduce another discriminator $D_{G}^{feat}$ further aligns feature distributions of $X_{N2G}$ and $X_G$ in the RGB domain:
\begin{footnotesize}
\begin{align}
\label{GAN_feature_G}
\nonumber & \mathcal{L}_{\text{GAN}}^{\text{feat}}\!\left(\!X_{\text{N}},X_{\text{G}},D_{\text{G}}^{\text{feat}},G_{\text{N2G}},F_{\text{G}}\!\right)\!\!=\!\mathbb{E}_{\text{x}_{\text{n}} \sim X_{\text{N}}}\!\left[D_{\text{G}}^{\text{feat}}\left(F_{\text{G}}(G_{\text{N2G}}(x_{\text{n}}))\!\right)\right] \\ 
& \quad \quad \quad +\mathbb{E}_{\text{x}_{\text{g}} \sim X_{\text{G}}}\left[D_{\text{G}}^{\text{feat}}\left(F_{\text{G}}(x_{\text{g}})\right)-1\right],
\end{align}
\end{footnotesize}

\textbf{Total Loss}
The full objective function in the fin-tuning stage is expressed as follows:
\begin{align}
\label{total_losses}
\mathcal{L} & =\mathcal{L}_{\text{pf}} +\lambda_{1}\mathcal{L}_{\text{tran}} +\lambda_{2}\mathcal{L}_{\text{GAN}}^{\text{feat}},
\end{align}
where $\lambda_{1}$ and $\lambda_{2}$ are hyperparameters, which we empirically set both as 1.

\section{Experimental Results}\label{section_results}

\begin{table}[t]\small
\begin{center}
\caption{Quantitative comparison among different NIR-to-RGB spectral domain translation methods.}
\centering
\label{table-1}
\resizebox{1\columnwidth}{!}{
\begin{tabular}{lcccc}
\toprule
\multirow{2}{*}{Methods}   & \multicolumn{4}{c}{Metrics} \\
& PSNR ($\uparrow$) & SSIM ($\uparrow$) & AE ($\downarrow$) & LPIPS ($\downarrow$) \\ 
\toprule
NIR-GNN\cite{9301839}~  & 17.50 & 0.60 & 5.22 & 0.384\\
MFF\cite{9301787}~   & 17.39 & 0.61 & 4.69 & 0.318 \\
SST\cite{9301788}~ & 14.26 & 0.57 & 5.61 & 0.361\\
ATCycleGAN\cite{9301791}~ & 19.59 & 0.59 & 4.33 & 0.295 \\
\textbf{MPFNet(Ours)}  & \textbf{22.14}  & \textbf{0.63} & \textbf{3.68} & \textbf{0.253} \\
\bottomrule
\end{tabular}
}
\end{center}
\end{table}

\subsection{Implementation and Training Details}
We used the VCIP2020 Grand Challenge on the NIR-to-RGB translation dataset for both training and testing. Specifically, there are 372 NIR-RGB image pairs in the training dataset and another 28 pairs for testing.  
We employed data augmentation by scaling, mirroring, random size cropping, and contrast adjustment.
Quantitative comparisons were performed using the Peak of Noise-to-Signal Ratio (PSNR), Angular Error (AE), Structural Similarity (SSIM)\cite{SSIM}, and LPIPS\cite{Zhang_2018_CVPR}.
All of the training images have the same size ($256\times256$) and are normalized to the range $(-1, 1)$. 
Firstly, we trained the domain translation module for 400 epochs with a learning rate $l_{\text{tran}} = 1 \times 10^{-4}$. Next, we trained our colorization network $F_\text{G}$ on $X_{\text{G}}$ for 250 epochs with a learning rate  $l_{\text{c}} = 1 \times 10^{-4}$. At last, we fine-tuned the whole network using the above pre-trained models. The batch size was set to 10.

\begin{figure}
    \centering
  { 
      \includegraphics[width=0.8\linewidth]{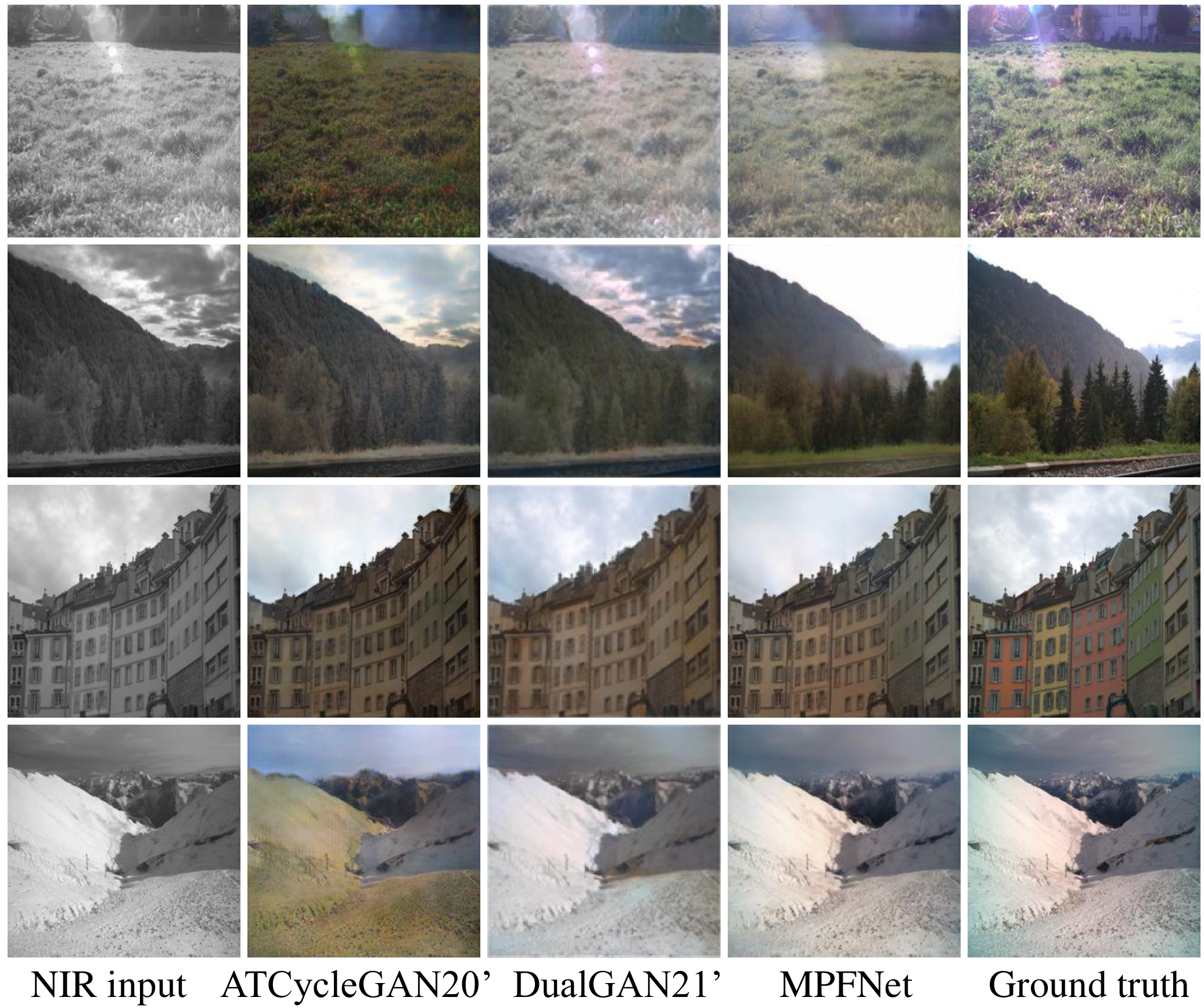}}
  \caption{Visual comparisons with ATCycleGAN\cite{9301791} and DualGAN\cite{liang2021improved} over the EPFL dataset. 
  }
    \label{comparison_EPFL} 
\end{figure}

\begin{table}[t]\footnotesize
\begin{center}
\caption{Comparisons on the EPFL dataset.}
\centering
\label{comparison_q_epfl} 
\resizebox{1\columnwidth}{!}{
\begin{tabular}{lcccc}
\toprule
\multirow{2}{*}{Methods}   & \multicolumn{3}{c}{Metrics} \\
& PSNR ($\uparrow$) & SSIM ($\uparrow$) & AE ($\downarrow$) & LPIPS ($\downarrow$) \\ 
\toprule
ATCycleGAN20'\cite{9301791}~  & 16.89 & 0.53 & 5.64 & 0.382 \\
DualGAN21'\cite{liang2021improved}~   & 17.80 & \textbf{0.62} & \textbackslash & \textbackslash \\
\textbf{MPFNet(Ours)}  & \textbf{20.61}  & 0.61 & \textbf{4.16} & \textbf{0.274} \\
\bottomrule
\end{tabular}
}
\end{center}
\end{table}

\subsection{Comparison with NIR Colorization Methods}

In this section, we quantitatively and qualitatively compare our MPFNet method with ATCycleGAN \cite{9301791}, NIR-GNN \cite{9301839}, MFF \cite{9301787}, and SST \cite{9301788}. 
As shown in Table \ref{table-1}, our method outperforms all existing methods.
Especially, compared to ATCycleGAN (1st Runner up of the VCIP 2020 Grand Challenge of NIR-to-RGB spectral domain translation) our method obtains a performance gain of \textbf{2.55} dB and \textbf{0.04} in terms of PSNR and SSIM, respectively.

For visual comparison, we randomly select five images from the test set and illustrate them in Fig. \ref{result_gragh_1}. As can be seen, our method can generate contextually-natural colorization results. The predicted results are closer to the style of ground-truth RGB images while retaining more texture information of NIR input images (\textit{e.g.}, the mountain in the third row).

To further validate the performance and generalization ability of our framework, we retrained both our network and ATCycleGAN \cite{9301791} using the EPFL dataset\cite{brown2011multi}, which has 477 NIR-RGB images pairs in total. Meanwhile, we also compared with the reported results of DualGAN\cite{liang2021improved}, which is also trained and evaluated on the same dataset. 
Note that the EPFL dataset has more complicated scene categories than the VCIP dataset, which involves urban, water, street, old buildings, and so forth. 
The quantitative and qualitative results of the test set are shown in Fig. \ref{comparison_EPFL} and Table \ref{comparison_q_epfl}, respectively. Obviously, our method still outperforms these two methods by a large margin.

\section{Acknowledgement}\label{acknowledgement}
This research is supported by A*STAR C222812026.

\section{Conclusion}\label{section_conclusion}
In this work, we have proposed a multi-scale progressive feature embedding framework for NIR-to-RGB spectral domain translation. Since the mapping relationship of the NIR-to-RGB translation is very implicit for models to capture, while the grayscale-to-RGB colorization is more explicit, we propose a domain translation module that translates NIR source images into the grayscale target domain, which significantly relieves the mapping ambiguity. To further improve the learning capacity, we have proposed a residual coordinate attention block to highlight objects of interest. Experiments show that our model achieves significant performance gains on the NIR-to-RGB spectral domain translation task.

\bibliographystyle{IEEEtran} 
\bibliography{mybib_nir}

\end{document}